\pdfoutput=1
\documentclass[11pt]{article}

\usepackage{acl}

\usepackage{times}
\usepackage{latexsym}

\usepackage[T1]{fontenc}
\usepackage[utf8]{inputenc}
\usepackage{graphicx}
\usepackage{cleveref}
\usepackage{booktabs}
\usepackage{multirow}
\usepackage[T1]{fontenc}
\usepackage[utf8]{inputenc}

\usepackage{microtype}

\title{Building Korean Sign Language Augmentation (KoSLA) Corpus  with Data Augmentation Technique}

\author{\\\textbf{Changnam An$^{1}$, Eunkyung Han$^{1}$, Dongmyeong Noh$^{1}$, Ohkyoon Kwon$^{1}$, Sumi Lee$^{1}$,} \\
  \textbf{ Hyunshim Han$^{2,3}$} \\
  \normalsize $^{1}$Yonsei University, Korea \\
  \normalsize $^{2}$KSL Interpreter (co-author)\\
  \normalsize $^{3}$Korea Nazarene University \\
  \texttt{\small \{iris12, topaz0828, dmnoh, kwon14, soom0916\}@yonsei.ac.kr, hanodilia@nate.com} \\
  } 

\begin{document}
\maketitle
\begin{abstract}
We present an efficient framework of corpus for sign language translation. Aided with a simple but dramatic data augmentation technique, our method converts text into annotated forms with minimum information loss. Sign languages are composed of manual signals, non-manual signals, and iconic features. According to professional sign language interpreters, non-manual signals such as facial expressions and gestures play an important role in conveying exact meaning. By considering the linguistic features of sign language, our proposed framework is a first and unique attempt to build a multimodal sign language augmentation corpus (hereinafter referred to as the KoSLA corpus) containing both manual and non-manual modalities. The corpus we built demonstrates confident results in the hospital context, showing improved performance with augmented datasets. To overcome data scarcity, we resorted to data augmentation techniques such as synonym replacement to boost the efficiency of our translation model and available data, while maintaining grammatical and semantic structures of sign language. For the experimental support, we verify the effectiveness of data augmentation technique and usefulness of our corpus by performing a translation task between normal sentences and sign language annotations on two tokenizers. The result was convincing, proving that the BLEU scores with the KoSLA corpus were significant.
\end{abstract}

\section{Introduction}

Sign languages are fully-fledged, complex, and natural languages with their own grammar, syntax, and vocabulary \cite{stokoe2005sign}. As in many countries, Korean sign language (KSL) is perceived as an independent language with the same qualifications as Korean. It is a language of its own linguistic system based solely on the visual and movement systems and the deaf culture. And like all other languages, it’s a living language that grows and changes over time\footnote{ \href{https://www.nad.org/resources/american-sign-language/what-is-american-sign-language/}{National Association of the Deaf}}.

There are roughly 70 million deaf people around the world and reportedly about 200 different sign languages\footnote{ \href{https://www.nad.org/resources/american-sign-language/what-is-american-sign-language/}{World Federation of the Deaf}}. Despite this situation, the research on sign language in the field of artificial intelligence is very insignificant compared to many other advances in the field \cite{yin2021including}. Many deaf people around the world and their languages have been relatively isolated from the recent progress of AI technology. To solve this problem, our study intends to contribute to creating a new corpus framework that will serve as a foundation for machine translation from general language to sign language.

A big difference between sign language and other spoken or written languages is that it is an ‘image-oriented language’. Sign languages use the visual-manual modality to convey meaning. We emphasize that the situations that occur due to the absence of hearing are common to the deaf. Thus, the need for related research and our proposed solutions are not limited to Korean sign language but can be universally applied to 200 other sign languages around the world that share the same challenges.

The purpose of this study is to demonstrate how to convert the general language, Korean in this paper, into annotated forms that contain the syntactic and semantic characteristics of the sign language used by the deaf. We suggest a universal framework for sign language augmentation corpus and neural machine translation.

The detailed scope of our project is to 1) build a sign language corpus for expressions frequently used and encountered by the deaf in hospitals, 2) overcome the issue of data scarcity through data augmentation without any structural and semantic changes, and ultimately 3) build a machine translation system that converts text into annotated sign language, which is the form just before visualization in sign language.

In this study, the conversation topic is limited to the hospital situation where the communication needs of the deaf are the highest, according to professional Korean sign language interpreters. It is to help doctors and deaf people communicate smoothly with each other in urgent and critical situations.

There are various fields of research on sign language. Research in each field has been actively conducted, such as mapping speech or text to sign language images or videos \cite{mesch2015gloss,fenlon2015building,gutierrez2016lse}, finding visual cues by sensing facial expressions, hand, and body movements of the deaf \cite{pishchulin2012articulated}, and making representations or symbols of sign language in units of meaning \cite{da2001signwriting}. Some of them have shown significant progress. 

Close to this study, there have been ongoing research projects that suggest a method of building a sign language corpus \cite{mesch2015gloss,johnston2010archive,konrad2018public}. However, there is a limit to the universal application of their sign language corpus, and above all, no case has tried to apply data augmentation technique to sign language and thoroughly consider the multimodal characteristics of sign language.

On the other hand, there are studies that directly convert text to visual sign language using neural networks \cite{stoll2020text2sign}. It is an attempt to make a latent representation by directly mapping a text input sequence to a motion graph consisting of a sequence of poses and gestures. In this case, however, there is a high possibility that the characteristics of sign language such as manual and non-manual signals and iconic features, will disappear during the translation process. In addition, if the linguistic features are not independently considered, there is a limitation in the ability to convey exact meanings due to the connotations of general languages. Therefore, ‘intermediate form’ is vital to make the meaning clear in sign language communication. The recent Korean sign language corpus provided by Korea’s AI Hub offers a mapping between words or sentences and their respective sign images but also lacks sufficient consideration of the linguistic features of sign language.

The need for an intermediate representation rather than a direct text-to-video translation in sign language is not the first to be raised in this study \cite{yin2020better}. According to their research, a new annotation scheme based on intermediate forms is necessary to minimize information loss in the translation process. Beyond shedding light on the issue, we provide a proper and concrete annotation framework, which is conducted in the same vein. Thus, this study focuses on building a corpus for sign language translation through data augmentation in the intermediate forms and applying the multimodality of sign language to it, to draw out the highly accurate results of machine translation between text to annotated sign language. 

Another notable issue in the field of AI research for sign language is to overcome data scarcity. Sign language is slightly different by region within a country, and it is hard to find any single, agreed-upon standard that is broadly applicable, which is also true in Korea. The quantity and quality of the dataset are critical to performance. Undoubtedly, poor data can lead to poor performance. To solve the problem of data scarcity, this study proposes and experimentally proves a data augmentation framework for sign language, which is reliable and universally applicable. For increasing the dataset reliability, we received support and advice from three professional sign language interpreters currently in the field.

Specifically, we tried to solve the data-side problem first by limiting the situation to hospitals and performing data augmentation through synonym replacement (SR) using the original dataset translated by sign language interpreters. It should be noted that the grammatical and semantic structures of sign language do not change during the augmentation process. This is how the KoSLA corpus was born, which can help enhance the efficiency and accuracy of sign language translation.

For the experimental support, several experiments were carried out on two tokenizers. The result was convincing, showing that the BLEU scores with the KoSLA corpus were significantly high. The two pillars of the KoSLA corpus are data augmentation and application of the multimodality of sign language. In this study, we focused on providing the experimental proof of data augmentation to increase the depth of the study. We reveal as a limitation that the application of multimodality remains as one of the future works.

We hope the KoSLA corpus serves as an excellent aid for the deaf while using medical services, and for those who want to learn sign language. We also urge more machine translation research considering the linguistic features of sign language in the future.

\section{Background and Basic Works}
Before explaining the corpus methodology and its structure defined for this study, we offer a brief introduction of the linguistic features of sign language. Then, we introduce the process of data collection and refinement that is aligned with the linguistic understandings, followed by our basic annotation rules of Korean-KSL translation.
\subsection{Linguistic Features of Sign Language}
Sign languages are largely composed of manual signals, non-manual signals, and iconic features (Figure 1). Linguistically, sign language consists of five cheremes\footnote{A basic unit of sign language; equiv. to a phoneme.} : handshape, location, movement, orientation, and non-manual component. The first four belong to the manual signal, which conveys meaning through the shape, position, angle, and movement of the hands. Non-manual signal is a gesture or action such as change of facial expression and body posture; the tilt, shake or nod of your head; and/or the hunching of one or both shoulders—that is used to add or specify meaning\footnote{\href{https://www.lifeprint.com/asl101/topics/nonmanual-signals.htm}{Bill Vicars on NMS at ASL University, 2012}} . Iconic features are a combination of several elements of manual and non-manual signals, which refers to the use of common forms or underlying images shared by sign language users to convey an intuitive meaning (National Institute of Korean Language, 2018).

\begin{figure}
    \centering
    \includegraphics[width=\linewidth]{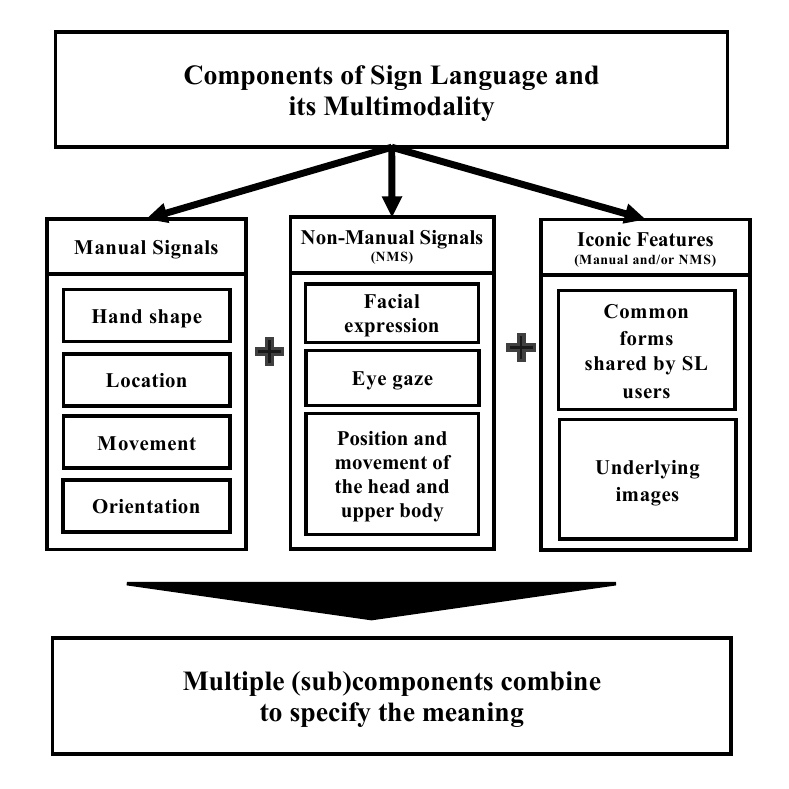}
    \vspace{-9mm}
    \caption{\textbf{Sign language components \& multimodality.} Sign language uses one or more components simultaneously to convey the meaning. Its visually focused multimodality distinguishes it from other spoken languages.}
    %\vspace{-5mm}
\end{figure}

It is necessary to pay attention to the importance of multimodality of sign language. Among the components mentioned above, the non-manual signals such as facial expressions and gestures also play an essential role in specifying meaning. However, in AI-based sign language research up to now, there have been almost no studies that properly reflected the linguistic features of sign language due to the difficulties in varying expressions and patterns of sign language. Most of the sign language studies that have been published so far are in the way of translating general grammar into sign language from the perspective of the Listener, or different from the actual sign language grammar and usage. This is true not only in linguistics research but also in artificial intelligence research. There is still a great need to fully understand linguistic features such as multimodality of sign language and apply it correctly to the field of AI.

To sum up, after considering the linguistic features of sign language, we suggest the KoSLA corpus, which is described as follows with its methodology and structures.

\newcommand{\STRUT}{\rule{0in}{2ex}}
\begin{table*}
\centering

\resizebox{\textwidth}{!}
{%
\begin{tabular}{l|l|l|l}
\toprule
\multicolumn{1}{c|}{\textbf{Category}} &
  \multicolumn{1}{c|}{\begin{tabular}[c]{@{}c@{}}\textbf{Original Sentence}\end{tabular}} &
  \multicolumn{1}{c|}{\begin{tabular}[c]{@{}c@{}}\textbf{Refined Sentence}\end{tabular}} &
  \multicolumn{1}{c}{\textbf{Notes}} \\ \hline
\multirow{2}{*}{Filtering} &
  \begin{tabular}[c]{@{}l@{}}I have a dry cough.\end{tabular} &
  \multirow{2}{*}{\begin{tabular}[c]{@{}l@{}}I have a cough.\end{tabular}} &
  \multirow{2}{*}{\textbf{\begin{tabular}[c]{@{}l@{}}Cannot distinguish the\\ coughing sound in sign language\end{tabular}}} \\ \cline{2-2}
 & \begin{tabular}[c]{@{}l@{}}I cleared my throat.\end{tabular} &              &  \\ \hline
\multirow{2}{*}{Paraphrasing} &
  \begin{tabular}[c]{@{}l@{}}Suffering from anemia.\end{tabular} &
  \begin{tabular}[c]{@{}l@{}}\textbf{I am} suffering from anemia.\end{tabular} &
  \multirow{2}{*}{\textbf{\begin{tabular}[c]{@{}l@{}}Supplements original sentence\\ considering sign language expressions\end{tabular}}} \\ \cline{2-3}
 & Desperate                                                      & Lack of hope &  \\ \bottomrule
\end{tabular}
}
\caption{\textbf{Korean refinement.} Sentences that cannot be expressed in sign language were deleted/replaced(filtering), and the original text was modified for parts that require clarity(paraphrasing).}
\end{table*}

\subsection{Dataset Collection \& Refinement}
The dataset for this study was limited to conversations in hospitals among various situations, based on interviews with professional sign language interpreters. In order to collect sentences that will be used in hospitals, 3,526 sentences from Korean hospitals were collected through AI Hub's hospital conversation (Korean-English) sentences and hospital conversation books.

We set refinement standards for collected sentences with the help of professional sign language interpreters (Table 1). The collected sentences were outlined by consulting them. First, sentences that cannot be expressed in sign language were removed or replaced by common sentences (filtering), and parts that require modification or simplification in subject, object, and context were appropriately changed in the original text (paraphrasing). As a result, a total of 610 sentences were selected.

% \usepackage{multirow}

\subsection{Korean-to-Annotated KSL Translation}
Refined Korean sentences were converted into annotated sign language through professional sign language interpreters. In this study, in order to apply the characteristics of sign language, manual signals, non-manual signals, and iconic features were separately defined in the learning data. Since sign language is an image language different from general language, Korean-KSL translation was performed as shown in Figure 2 to minimize information loss by taking into account the multimodality of sign language.
\begin{figure}
    \vspace{-3mm}
    \includegraphics[width=\linewidth]{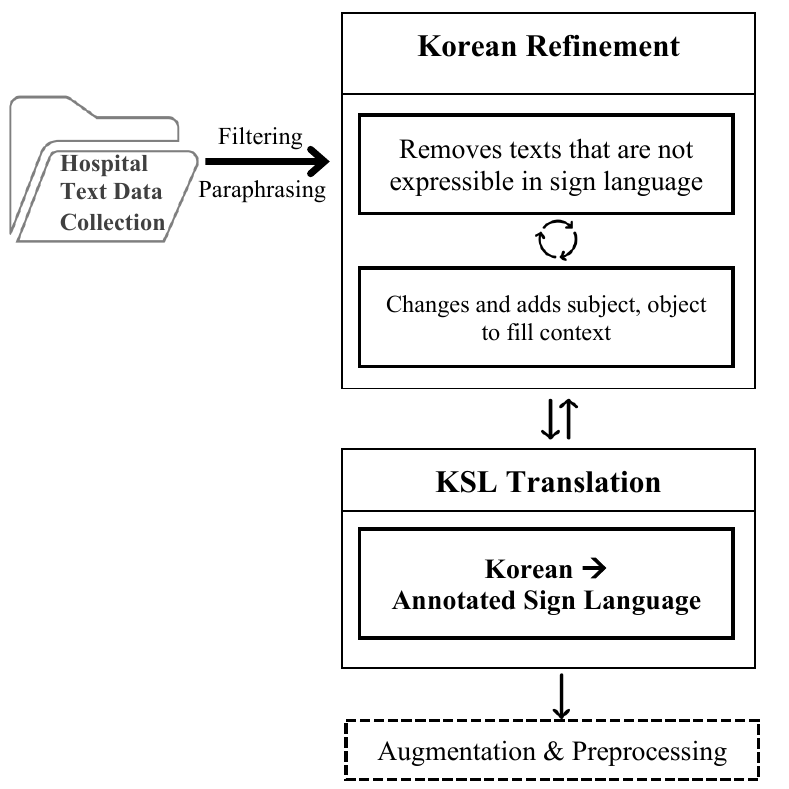}
    \caption{\textbf{Dataset collection, Korean refinement and KSL(Korean Sign Language) translation.}  Translation was performed by interpreters. Sentences irrelevant from the deaf were removed after the sentences were collected from the hospital context, and original Korean sentences were refined for translation into sign language annotations.}
    \vspace{-4mm}
\end{figure}

In general, machine translation uses a one-to-one mapping. But our learning data translated into annotated sign language from Korean was divided into manual signals, non-manual signals, and iconic features. That is, the data was defined separately so that it can be learned later in a multilayer form. This can be seen as a differentiating factor.

In Table 2, the sentence "Are you nauseous?" is translated into sign language annotation as “(vomiting action) want?[BE]”, with BE representing as bigger eyes. If the annotated sign language is used as a source without abbreviation, information loss may occur, in which the iconic feature corresponding to (vomiting action) is changed unintentionally to a non-manual signal. Therefore, we separated the information of iconic features and non-manual signals in the source sentence and used them in each layer form.

\begin{table}[]
\vspace{-3mm}

\begin{tabular}{cc}
\toprule
\multicolumn{1}{c|}{\textbf{Mark}} & \textbf{Definition}       \\ \hline
\multicolumn{1}{c|}{( )}           & Iconic features        \\ \hline
\multicolumn{1}{c|}{{[} {]}}       & Non-manual signals     \\ \hline
\multicolumn{1}{c|}{/}             & Separator of each element \\ \hline
\multicolumn{2}{l}{\begin{tabular}[c]{@{}l@{}}Example.\\Spoken language sentence: Are you nauseous?\\       → Annotated Sign language:\\ \hspace{5mm}(vomiting action) / want?{[}BE{]}\\  \hspace{4mm} 
*BE: Bigger Eyes\end{tabular}} \\ \bottomrule
\end{tabular}
%\vspace{-6mm}
\caption{\textbf{Prevention of information loss through additional marking using parentheses and brackets.} “BE” is an abbreviation of  ‘Bigger Eyes’, which is an example of applying shortening rules for improving performance by alleviating the complexities of long expressions.}
\end{table}

In an effort to obtain accurate results, we applied a translation guide so that 1) each expression of iconography/non-manual signal was recognized as a single element to avoid being perceived as a sentence, 2) the expressions of each action and emotion were abbreviated for simplification, and 3) complex sentences were made simpler in a sentence.

\section{The KoSLA (Korean Sign Language Augmentation) Corpus}
In order to improve the performance of machine translation, we performed data augmentation and classification of manual signals, non-manual signals and iconic features (Figure 3).
\subsection{Data Augmentation by Keywords}
Machine translation requires a big amount of learning data. Initial experiments with insufficient translation datasets did not show noteworthy results. Instead of more resource-intensive dataset building, we used data augmentation techniques to overcome the limitations in time and cost.

Due to the inadequate amount of sign language data and the structural complexity of sign language grammar, it was difficult to use techniques such as UDA \cite{xie2019unsupervised} and back-translation \cite{sennrich-etal-2016-improving}, which are currently in wide use. In addition, the Example Interpolation technique \cite{zhang2017mixup,guo2020sequence}, which is being actively studied in the recent years, and the DA technique  using the pre-trained model \cite{kumar2020data,kobayashi2018contextual,yang2020generative} were also excluded from considerations for the same reason.

In choosing a data augmentation technique, we focused on the point that the most ideal DA technique must be easy to implement, preserve its own meaning well, and improve the performance of the model. Among them, rule-based DA had the advantage of being easy to implement and providing gradual performance improvement \cite{li2017robust,wei2019eda,wei2021text}. We selected it as the most suitable DA technique for our research.

\begin{figure}
    \centering
    \includegraphics[width=\linewidth]{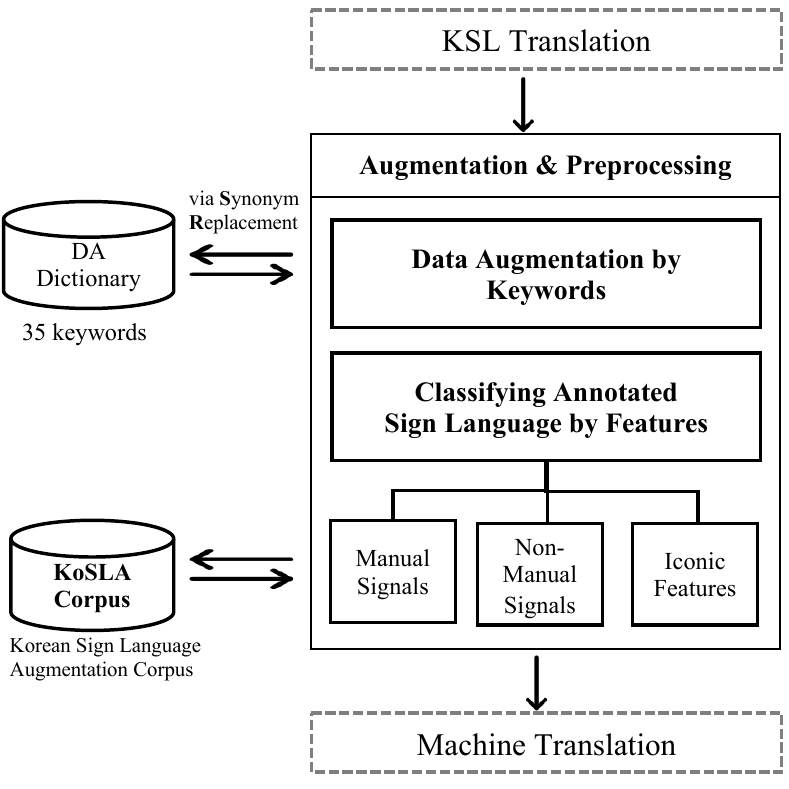}
    \vspace{-7mm}
    \caption{\textbf{Data augmentation \& preprocessing.}
Data augmentation is performed by using keywords and sign language annotations are classified by marks of features.}
\end{figure}

Among the rule-based techniques \cite{feng2021survey}, we used the SR technique among the four techniques presented in the paper "EDA: Easy Data Augmentation Techniques” (Table 3). With this technique, the structure of the annotated dataset was maintained, minimizing the information loss.
\begin{table}[]
%\vspace{1mm}

\renewcommand{\tabcolsep}{0.7mm}
\begin{tabular}{l|l}
\toprule
\multicolumn{1}{c|}{\textbf{Technique}} &
  \multicolumn{1}{c}{\textbf{Description}} \\ \hline
\begin{tabular}[c]{@{}l@{}}Synonym\\ Replacement (SR)\end{tabular} &
  \begin{tabular}[c]{@{}l@{}}A word selected randomly\\is replaced with its synonym\\in the same position.\end{tabular} \\ \hline
\begin{tabular}[c]{@{}l@{}}Random\\ Insertion (RI)\end{tabular} &
  \begin{tabular}[c]{@{}l@{}}A word selected randomly\\is inserted with its synonym\\in a random position.\end{tabular} \\ \hline
\begin{tabular}[c]{@{}l@{}}Random\\ Swap (RS)\end{tabular} &
  \begin{tabular}[c]{@{}l@{}}Positions of two words\\selected randomly\\are swapped.\end{tabular} \\ \hline
\begin{tabular}[c]{@{}l@{}}Random\\ Deletion (RD)\end{tabular} &
  \begin{tabular}[c]{@{}l@{}}Each word in the sentence\\is deleted with probability p.\end{tabular} \\ \bottomrule
\end{tabular}
\caption{\textbf{Easy data augmentation techniques} \cite{wei2019eda}}
%\vspace{-6mm}
\end{table}
Simply put, we added special symbols (e.g., [a\_Disease name\_a]) to each sentence for synonym replacement. 1) The DA dictionary is made for replaceable category names, and 2) a new sentence is generated by replacing each word belonging to the dictionary category with the special symbol. For dictionary building, we used the Korean Standard Classification of Diseases (KCD) on the International Classification of Diseases (ICD-10), and the facility information registered on the Yonsei Severance Hospital website. As a result of this data augmentation, the total number of sentences became 40,558, which is about 67 times larger compared to the original dataset.
\subsection{Classifying Annotated Sign Language by Features}
First, we performed classification by extracting the iconic features and non-manual signals in the annotated sign language translated by professional sign language interpreters, and maintaining the location information of each sentence at its matching position in the expression. Sign language expressions in the feature classification step are divided into iconic features and non-manual signals (Figure 4).
\\
In order to distinguish between iconic features and non-manual signals, sentences were separated by inserting \{ICON\} at the positions of iconic features and [NMS] at the positions of non-manual signals (Figure 5). The early corpus built this way went through another data augmentation process.
\begin{figure}
    %\vspace{-24mm}
    %\centering
    \includegraphics[width=\linewidth]{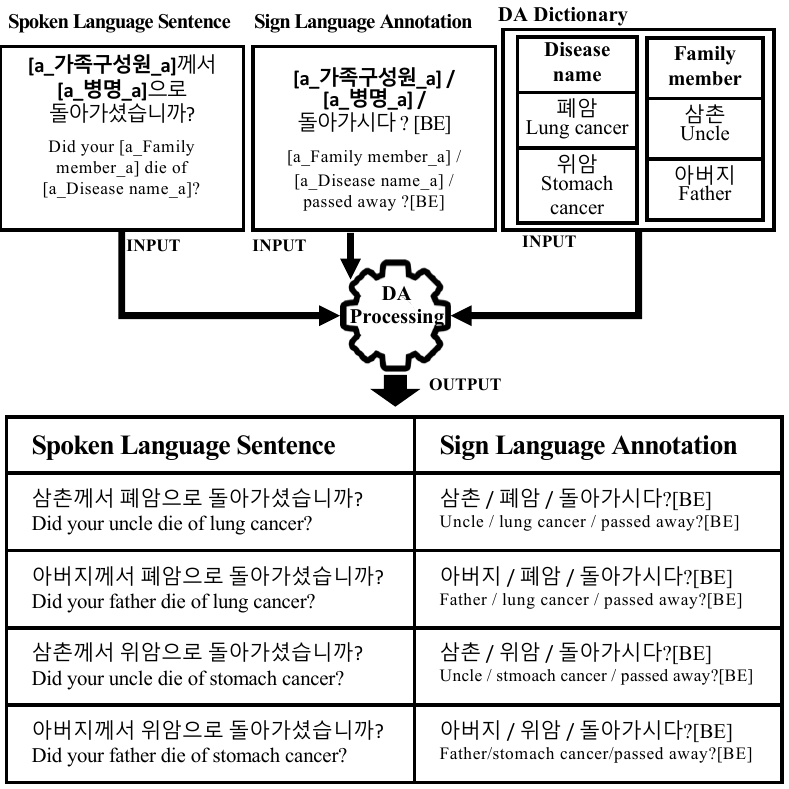}
    \vspace{-7mm}
    \caption{\textbf{Data augmentation using DA dictionary} \cite{wei2019eda}. 
It shows the generation of 4 pairs of sentences of spoken language and sign language. Words that include [a\_Family member\_a] and [a\_Disease name\_a] are mapped to the corresponding category of the DA dictionary.}
    %\vspace{-4mm}
\end{figure}

\section{Model Design}
Before applying the DA technique and the translation dataset to the model, we reviewed structural similarities and differences of Korean language and Korean sign language.
\subsection{Applying Tokenizer}
In general, a tokenizer is required for each language for machine translation between two languages (e.g., Korean-English). However, in the case of Korean-KSL, the same tokenizer can be used because annotated sign language is based on Korean language. To generate the tokenizer's vocabulary, we used colloquial Korean sentences provided by AI Hub.
\subsection{Necessity of Multilayer Corpus}
Prior to designing the translation model, it is important to make the model understand the context of the input sentence. For this, we reasoned that a dataset consisting of sentences with similar contexts would be needed. Learning models and experiments were focused on verifying whether the DA-assisted model was good at translating by recognizing context patterns.

\begin{figure}
    %\vspace{-53mm}
    %\centering
    \includegraphics[width=\linewidth]{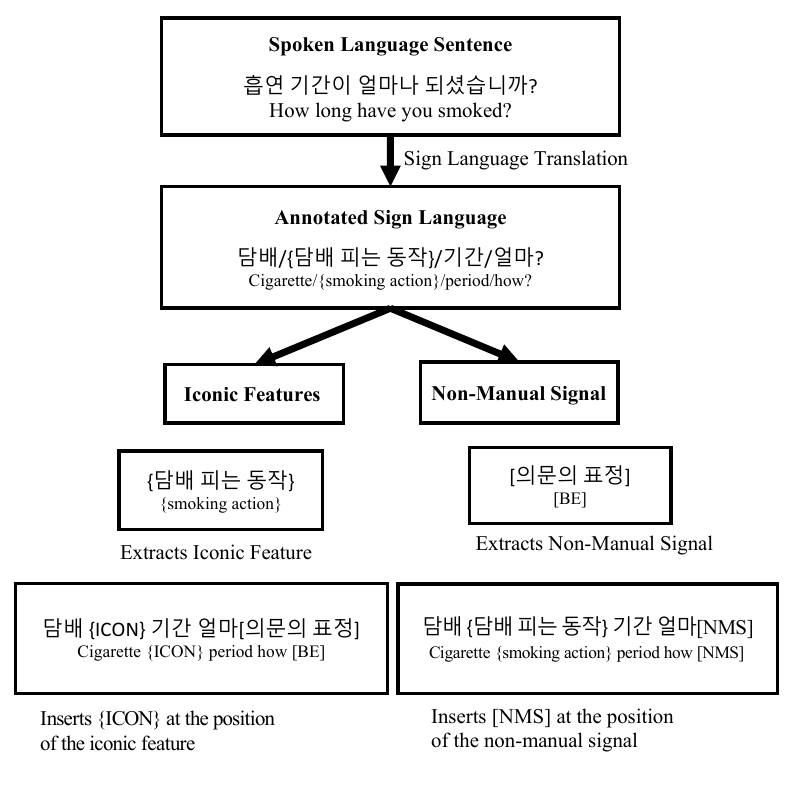}
    \vspace{-8mm}
    \caption{\textbf{Per-expression separation of sign language annotations.} Non-manual signals and iconic features are marked with special symbols, after hospital sentences are converted to sign language annotations. Non-manual signals and iconic features are extracted afterwards.}
    \vspace{-4mm}
\end{figure}

Other research on sign language translation including Turkish and Chinese used various models such as Sequence to Sequence, LSTM, and RNN \cite{kayahan2019hybrid,arvanitis2019translation,guo2018hierarchical,gao2021rnn,mesch2015gloss,gutierrez2016lse}.
Our research used Transformer \cite{vaswani2017attention} as it basically shows high performance on natural language processing.

We referred to the encoder and decoder part of Transformer, which handles a method of simultaneously performing voice recognition and translation \cite{le2020dual} and a method of voice recognition where two languages are simultaneously used \cite{zhou2020multi}. In order to devise methods for multimodal learning, we were also inspired by the ideas of using video, audio, and text in a similar way to how sign language expressions are divided into manual signals, non-manual signals and iconic features \cite{akbari2021vatt} and finding keywords in video and audio data \cite{tsai2019multimodal}. In the experimental step, we centered on the method of inputting the data of manual signals, non-manual signals, and iconic features into the model in a multilayer form.

\begin{figure}
    \centering
    \includegraphics[width=\linewidth]{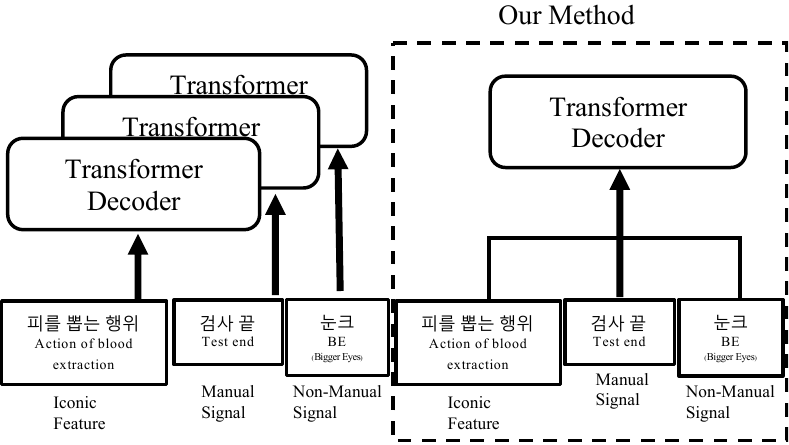}
    \caption{\textbf{Giving manual signals, non-manual signals, and iconic features to Transformer’s decoder as input.} The left side shows adding Transformer’s multiple decoders. The right side shows translation performed with a single decoder (our method).}
    \vspace{-4mm}
\end{figure}
First, an experiment on separating word segments was conducted in which sign language annotations matching Korean sentences were inputted to the model without multimodal applications. Translation results were not satisfactory for initial experiments in which manual signals, non-manual signals, and iconic features were not seperated, showing outputs completely different from the desired ones. For example, the sign language expression corresponding to "blood test" is "(blood extraction action)/test" in our annotations. If we separate the expression simply by its segments, information loss is inevitable as every element of the expression is grouped into one sentence like "blood extraction action test”, losing its structural meaning.

Multimodal information on manual signals, non-manual signals, and iconic features was required to solve such problems, and we also gave it the name of ‘multilayer’ in the model design.

We reviewed two ways of constructing multilayer: modifying Transformer’s decoder module and configuring input data. The experiment was conducted in the right method of Figure 6, using a single decoder to reduce the complexity of the model.
Figure 7 shows the detailed method for Korean-KSL translation.

We verified the quality of translation in this study by using the manual signals, leaving verification with non-manual signals and iconic features as a future work. That is, only the part of manual signals in Figure 7 was used for the quality support.

\begin{figure}
    \centering
    \includegraphics[width=\linewidth]{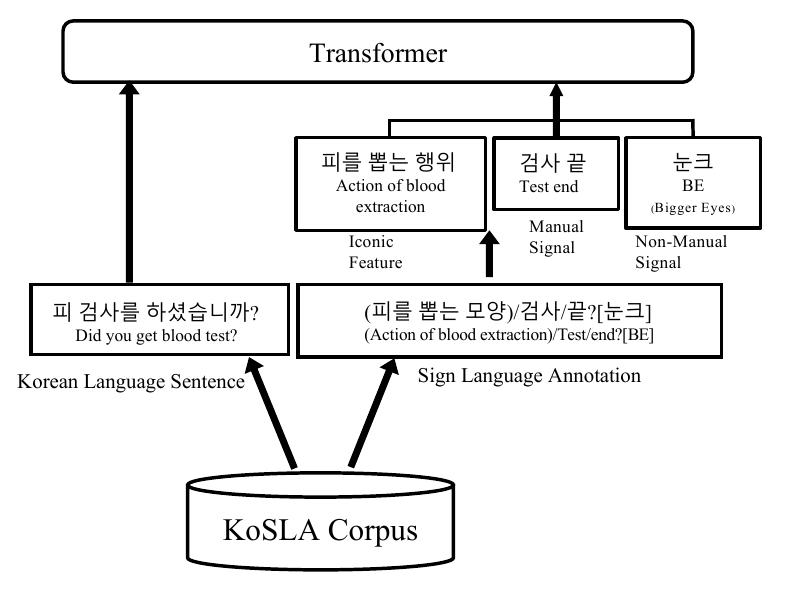}
    \vspace{-7mm}
    \caption{\textbf{Inputting model after separating data of manual signals, non-manual signals and iconic features of sign language annotations}}
    \vspace{-2mm}
\end{figure}

The early experiment was conducted only with a non-DA translation dataset, and the translation was a failure. We came to the conclusion that the dataset was not large enough for the model to recognize the structure of the sentence. To find an answer to this, we used the DA technique in the experiment and it showed a better performance than the previous test run. However, the translation performance was still not satisfactory when the context of the original Korean sentence became slightly different.

To overcome the problem, we reviewed various tokenizer methods of segmenting Korean sentences \cite{eo2021research}. We decided to use the LTokenizer of soynlp as the base model in this study. Experiments showed that the use of SentencePiece performs better in solving the Out of Vocabulary(OOV) problem.
Table 4 shows that by using SentencePiece, translation is performed quite accurately even when the context is changed.

\begin{table}[h]
\vspace{1mm}

\begin{tabular}{l|l}
\toprule
\multicolumn{1}{c|}{\begin{tabular}[c]{@{}c@{}}\textbf{\small Input Sentence}\end{tabular}} &
  \multicolumn{1}{c}{\begin{tabular}[c]{@{}c@{}}\textbf{\small Translated Sentence}\end{tabular}} \\ \hline
\begin{tabular}[c]{@{}l@{}}Is\_[normal expression]\\my sister-in-law suffering\\from kidney disease?\end{tabular} &
  \begin{tabular}[c]{@{}l@{}}sister-in-law\\ kidney disease\quad\\ in progress\end{tabular} \\ \hline
\begin{tabular}[c]{@{}l@{}}Is\_[honorific expression]\\my brother suffering\\from kidney disease?\end{tabular} &
  \begin{tabular}[c]{@{}l@{}}brother\\ kidney disease\\ in progress\end{tabular} \\ \bottomrule
\end{tabular}
%\vspace{2mm}
\caption{\textbf{Translation result after applying SentencePiece tokenizer.} Result of translating Korean sentences which have differences in subject(sister-in-law → brother) and verb(is\_[normal expression] → is\_[honorific expression]) into annotated sign language.}
\vspace{-1mm}
\end{table}

Table 5 shows the result of using two datasets, DA and Non-DA, when applying soynlp and SentencePiece tokenizers. The data was augmented about 67 times by applying 35 DA dictionary categories using the SR technique.
\begin{table}[]
\vspace{-1mm}

\begin{tabular}{|cr|cr}
\toprule
\multicolumn{2}{c|}{\textbf{Exp. 1 DA}} & \multicolumn{2}{c}{\textbf{\quad\quad Exp. 2 Non-DA\quad\quad }} \\ \hline
\multicolumn{1}{c|}{\quad Train \quad\quad}   & 28,390   & \multicolumn{1}{c|}{\quad Train \quad\quad}       & 427      \\ \hline
\multicolumn{1}{c|}{\quad Test \quad\quad}    & 4,057    & \multicolumn{1}{c|}{\quad Test \quad\quad}        & 61       \\ \hline
\multicolumn{1}{c|}{\quad Valid \quad\quad}   & 8,111    & \multicolumn{1}{c|}{\quad Valid \quad\quad}       & 122      \\ \hline
\multicolumn{1}{c|}{\quad Total \quad\quad}   & 40,558   & \multicolumn{1}{c|}{\quad Total\quad\quad}       & 610      \\ \bottomrule
\end{tabular}
%\vspace{4mm}
\caption{\textbf{Number of DA and Non-DA datasets for experiments.}}
\end{table}
%\newpage

%\input{acl-ijcnlp2021-Hear/5.Experiment_Results}
\section{Experiment Results}
The KoSLA-based method showed significantly high performance on both tokenizers (Table 6). The BLEU score of SentencePiece is proven better for Korean-KSL translation than that of soynlp (Table 7).
LTokenizer of soynlp had an OOV problem by separating the Korean word segments into noun + particle form. The BLEU score of soynlp was high at 91.93, but when digging into the segmented words, we could see that the score was high because lots of words were processed as <unk> and then translated into the same <unk> by the model. On the other hand, SentencePiece showed better performance because it learns directly without pre-split process. In addition, SentencePiece's translation results did not show many <unk> tokens.

\section{Conclusion}
This paper proposes the KoSLA corpus to solve the problem of data scarcity, which poses as one of the hurdles for artificial intelligence learning. We also tried to improve inaccurate translation that does not reflect the linguistic characteristics of sign language. We built a sign language corpus with data augmentation technique like synonym replacement. And expressions of manual signals, non-manual signals, and iconic features were classified and defined in multimodality to minimize information loss of sign language in the translation process. Applying this method to the Transformer model resulted in remarkable performance improvement. We are planning to continue our research on various DA techniques and translation models for sign language. We hope the KoSLA corpus can lead to more AI research focusing on linguistic features of sign language such as multimodality and serve as a first aid for the deaf when using medical services and for those who want to learn sign language.
\section*{Acknowledgments}
We would like to thank Professor Hong-Goo Kang of the Department of Electrical and Electronic Engineering at Yonsei University, for providing insightful feedback during research planning and related review. In addition, we would also like to express our deepest gratitude to professional sign language interpreters Jeung-eun You, Hyeon-ya Oh and Myung-hee Kim for their great help in improving the quality and quantity of the sign language translation dataset.
\begin{table}[]
%\vspace{-40mm}
\caption{\textbf{Result of train/validation loss by tokenizers}}
\renewcommand{\tabcolsep}{0.7mm}
\begin{tabular}{l|cr|cr}
\toprule
\multirow{2}{*}{\textbf{\quad Tokenizer \quad}} &
  \multicolumn{2}{c|}{\textbf{\begin{tabular}[c]{@{}c@{}}Train\\ Loss\end{tabular}}} &
  \multicolumn{2}{c}{\textbf{\begin{tabular}[c]{@{}c@{}}Validation\\ Loss\end{tabular}}} \\ \cline{2-5} 
 &
  \multicolumn{1}{c|}{\textbf{\begin{tabular}[c]{@{}c@{}}DA\end{tabular}}} &
  \multicolumn{1}{c|}{\textbf{\begin{tabular}[c]{@{}c@{}}Non-DA\end{tabular}}} &
  \multicolumn{1}{c|}{\textbf{\begin{tabular}[c]{@{}c@{}}DA\end{tabular}}} &
  \multicolumn{1}{c}{\textbf{\begin{tabular}[c]{@{}c@{}}Non-DA\end{tabular}}} \\ \hline
\begin{tabular}[c]{@{}l@{}}soynlp\\ Ltokenizer\\ (baseline)\end{tabular} &
  \multicolumn{1}{r|}{0.0858} &
  0.2373 &
  \multicolumn{1}{r|}{\textbf{0.0882}} &
  12.47 \\ \hline
SentencePiece & \multicolumn{1}{r|}{0.0301} & 0.5210 & \multicolumn{1}{r|}{\textbf{0.0927}} & 12.79 \\ \bottomrule
\end{tabular}
\end{table}

\begin{table}[]
\vspace{3mm}
\caption{\textbf{Result of BLEU scores by tokenizers.} The application of SentencePiece tokenizer without pre-split process shows better performance than soynlp.}
\begin{tabular}{l|r}
\toprule
\multicolumn{1}{c|}{\textbf{\begin{tabular}[c]{@{}c@{}}Tokenizer\end{tabular}}} &
  \multicolumn{1}{c}{\textbf{\begin{tabular}[c]{@{}c@{}}\quad BLEU Score\quad \end{tabular}}} \\ \hline
\begin{tabular}[c]{@{}l@{}}soynlp Ltokenizer (baseline)\end{tabular} &
  \begin{tabular}[r]{@{}l@{}}91.93\end{tabular} \\ \hline
\begin{tabular}[c]{@{}l@{}}SentencePiece\end{tabular} &
  \begin{tabular}[r]{@{}l@{}}\textbf{95.82}\end{tabular} \\ \bottomrule
\end{tabular}
\end{table}
\newpage

%\bibliography{anthology,reference}
\bibliography{reference}

\end{document}